\documentclass[10pt, a4paper]{article}
\usepackage{tabularx}
\usepackage{amsmath}
\usepackage[final]{lrec2026} 

\title{MasonNLP at MEDIQA-SYNUR 2026: Retrieval-Augmented Large Language Models for Schema-Constrained Clinical Information Extraction}

\name{A H M Rezaul Karim, \"Ozlem Uzuner} 

\address{George Mason University, VA, USA \\
         akarim9@gmu.edu, ouzuner@gmu.edu \\}

\abstract{
Conversational nurse-patient transcripts contain actionable observations, but converting these transcripts into structured representations at scale remains challenging. Documentation burden is substantial, with prior studies showing clinicians spend large portions of their workday on documentation and related desk work rather than direct patient care. MEDIQA-SYNUR focuses on observation extraction from conversational nurse-patient transcripts, requiring systems to normalize these narratives into a predefined schema with value-type constraints. We propose a modular retrieval-augmented generation (RAG) pipeline that uses the training set as an exemplar corpus, combines schema-constrained prompting (full schema vs.\ pruned candidate schema), deterministic schema-based postprocessing, and a second-pass audit, with two LLM backbones: \textit{Llama-4-Scout-17B-16E-Instruct} and \textit{GPT-5.2} with corresponding embedding models for RAG. Our best configuration uses \textit{GPT-5.2} with full schema, RAG, and a second-pass auditing, achieving 80.36\% $F_1$ score. Overall, our results show that RAG consistently improves performance, while the optimal degree of schema constraint depends on the model, and second-pass auditing yields modest additional gains by correcting residual schema-adherence errors.\newline \Keywords{Retrieval Augmented Generation (RAG), Clinical Information Extraction, LLM}}

\begin{document}

\maketitleabstract
\section{Introduction}
Clinical documentation is essential for continuity of care and quality reporting, yet it remains a major source of clinician workload and professional dissatisfaction \cite{muhiyaddin2022electronic}. A study in ambulatory practice has shown that physicians spend nearly half of their clinic day on electronic health record (EHR) and desk work (49.2\%) and substantially less time in direct clinical face time (27.0\%), with additional after-hours work of approximately 1-2 hours per night devoted mostly to EHR tasks \citep{sinsky2016allocation}. Similar EHR log-based analyses report that primary care clinicians spend 145.9 minutes/day actively using the EHR \citep{rotenstein2022association}. These high demands for documentation motivate methods that can automatically capture and convert clinical information into structured representations with minimal manual entry. As a step in this direction, particularly for workflows such as nursing assessments, clinical documentation is frequently transcribed~\cite{mayer2022comparison}. Conversational nurse-patient transcripts are examples of such documentation. 


Information extraction (IE) can convert transcripts into structured representations \cite{balasubramanian2025leveraging,hu2026information}.  In many clinical IE settings, the goal is to identify spans, assign concept labels, or extract relations from clinical text, often within relatively limited label spaces and without requiring the model to generate a fully normalized structured output. Strict requirements on output, such as conforming to a large predefined schema with consistent formatting and validity constraints \citep{karim-uzuner-2025-masonnlp, builtjes2025leveraging} pose additional challenges for clinical IE. 

MEDIQA-SYNUR task presents an observation extraction task that encompasses these challenges, with a specific focus on conversational nurse-patient transcripts. Given a transcript, MEDIQA-SYNUR requires automated systems to extract clinically salient observations normalized to a predefined schema with explicit value-type requirements (e.g., numeric vs.\ categorical values and enumerated option validity) \citep{mediqa-synur-task}. The \emph{schema} for the task captures the complete set of target observation concepts together with their value types and any allowable categorical value sets. The SYNUR dataset accompanying this task provides an open-source synthetic corpus annotated by expert nurses with a large set of structured observations, enabling controlled evaluation of observation extraction from conversational nurse-patient transcripts \citep{mediqa-synur-dataset}. 

Recent advances in instruction-following large language models (LLMs) have enabled prompt-based clinical IE for converting clinical narratives into structured representations \citep{agrawal-etal-2022-large, rodrigues2025harnessing}.  Compared with task-specific supervised systems, LLMs offer a more flexible framework, and prior work has shown that, with carefully designed prompts, they can support few-shot extraction across clinical IE tasks \citep{agrawal-etal-2022-large, rodrigues2025harnessing}. However, reliability and schema adherence remain persistent concerns, particularly as output spaces grow. LLMs tend to hallucinate concept names, violate type constraints, and produce inconsistent formatting when schema complexity increases \citep{karim-uzuner-2025-masonnlp,builtjes2025leveraging}. Evidence also suggests that open-weight and closed-weight models exhibit meaningfully different instruction-following behaviors and robustness profiles under constraint-heavy prompts \citep{builtjes2025leveraging}.  In this work, we explore open-weight and closed-weight LLMs in observation extraction by constraining their output through a schema that varies in its complexity.  We refer to this as schema-constrained observation extraction.

We hypothesize that retrieval-augmented generation (RAG) \cite{lewis2020retrieval} can improve performance in observation extraction by conditioning output on retrieved exemplars \citep{shlyk2024real,lopez2025clinical,zhan2025ramie,liu2025improving}. 
To test this hypothesis, we study two RAG \citep{lewis2020retrieval} approaches that use the training set as the retrieval corpus. For \textit{Llama-4-Scout-17B-16E-Instruct} \citep{meta2025llama}, we find that constraining the output space using a pruned candidate schema improves results. In contrast, with \textit{GPT-5.2} \citep{openai_gpt52_2025}, RAG yields the best performance with full schema, and a second-pass auditing provides additional gains.\footnote{Implementation details can be found here: https://github.com/AHMRezaul/MEDIQA-SYNUR-2026}

\begin{figure*}[!ht]
\begin{center}
\includegraphics[width=\textwidth]{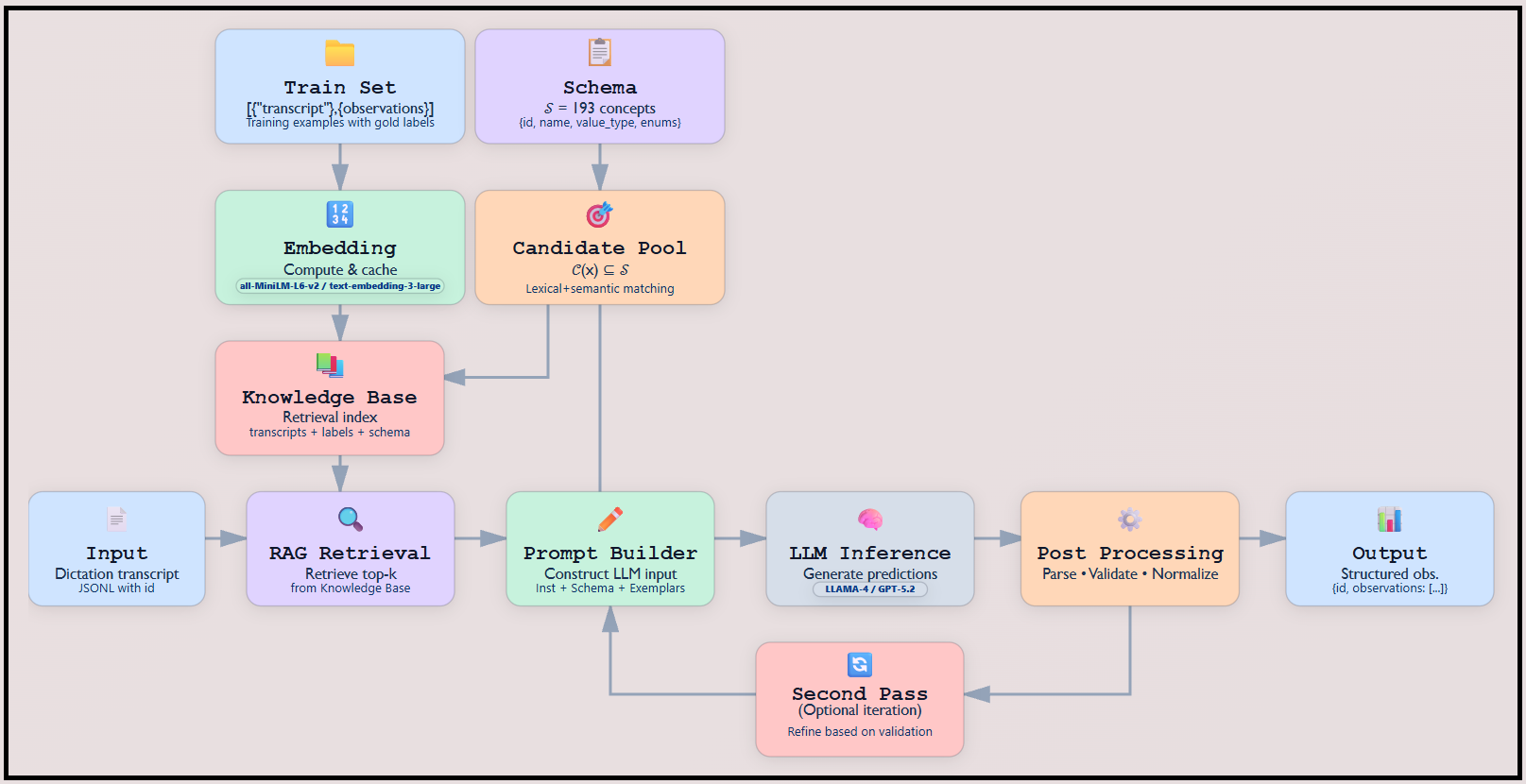}
\caption{Retrieval-augmented, schema-constrained pipeline that retrieves training exemplars, conditions the LLM on schema (full or pruned candidate), and post-processes outputs with second-pass auditing.}
\label{fig.1}
\end{center}
\end{figure*}

\noindent Our study makes the following contributions:
\begin{itemize}
\item The question of the interaction of RAG with the schema is in Clinical IE.  Our work directly addresses this gap by studying how RAG interacts with different schema constraints across LLM backbones of differing parameter scales, providing a systematic analysis of this interaction in observation extraction.
\item We show that a pruned candidate schema is beneficial for a smaller open-weight model but counterproductive for a larger model, suggesting that the formulation of the output constraints, namely how the fixed task schema is presented to the model during generation, should be model-aware, i.e., adapted to the behavior and capabilities of the underlying model.
\item We present second-pass auditing that provides modest gains primarily by correcting schema-adherence and normalization errors, but does not substitute for RAG and schema constraint, indicating it is best used as a final refinement stage rather than a core component.
\end{itemize}

\noindent Overall, our findings provide practical evidence that schema-constrained observation extraction should be model-aware. Our results clarify when a pruned candidate schema is helpful versus when a full schema is preferable, and further show how RAG and second-pass auditing can be combined to improve robustness.

\section{Related Work}

Clinical information extraction (IE) has been driven by tasks that standardize evaluation for concept extraction, assertion detection, and relations \citep{uzuner20112010,henry20202018,fu2020clinical,navarro2023clinical}. Benchmarks such as i2b2/VA and n2c2 established protocols for span extraction and normalization, but largely focus on relatively small label spaces \citep{uzuner20112010,henry20202018,mahajan2023overview}. In contrast, MEDIQA-SYNUR focuses on a large, heterogeneous output space (193 typed observation concepts), which must be extracted under strict type and enumeration constraints, motivating methods that explicitly condition generation on schema.

Early studies of conversational nurse-patient transcripts coupled speech recognition with IE to produce structured handover documents, highlighting both feasibility and the difficulty of extracting structured representations from transcripts \citep{johnson2014systematic,johnson2014comparing,dawson2014usability}. Subsequent shared evaluations released transcribed handover datasets and structured annotations, often using synthetic data to mitigate privacy barriers \citep{suominen2015benchmarking,suominen2015capturing,suominen2016task}. Building on this line, \citet{mediqa-synur-dataset} introduced SYNUR to support systematic evaluation of structured output from conversational nurse-patient transcripts, leaving open how to enforce schema adherence reliably at scale.

Within LLM-based clinical IE, most prior work has emphasized direct prompting or few-shot extraction, often in settings where outputs are interpreted more flexibly and are not tightly constrained by a large predefined schema \citep{agrawal-etal-2022-large, rodrigues2025harnessing}. Work that explicitly examines schema adherence has shown that structured extraction becomes substantially more difficult as output spaces grow and validity constraints become stricter \citep{karim-uzuner-2025-masonnlp,builtjes2025leveraging}. Although retrieval-augmented generation has begun to show promise in clinical extraction by grounding predictions in relevant examples or evidence \citep{lewis2020retrieval,shlyk2024real,lopez2025clinical,zhan2025ramie,liu2025improving}, its interaction with schema-constrained generation remains underexplored. Likewise, limited prior work has compared how open-weight and closed-weight LLMs behave under such constraint-heavy extraction settings. Our work addresses these gaps in the context of schema-constrained observation extraction.

\section{Task Description}
\subsection{Problem Formulation}
MEDIQA-SYNUR defines the task as an \emph{observation extraction} from conversational nurse-patient transcripts. Given a transcript $x$, the goal is to identify clinically salient observations and normalize them to a predefined schema.

The schema is a set of $M=193$ observation concepts $\mathcal{S}=\{c_1,\ldots,c_M\}$. Each concept $c_m$ has an identifier $\mathrm{id}_m$, a name $N_m$, a value type $\tau_m \in \{\textsc{single\_select}, \textsc{multi\_select}, \textsc{numeric}, \textsc{string}\}$, and, for categorical types, an allowed value set $\mathcal{V}_m$.

For an input $x$, a system outputs a list of extracted observation instances
$\hat{O} = [o_1,\ldots,o_n]$, where each instance is an object from the schema $o_i=\{\hat{\mathrm{id}}_i,\hat{N}_i,\hat{\tau}_i,\hat{v}_i\}.$
The evaluation measures the correctness of the predicted observations under this schema.

\subsection{Dataset}
The SYNUR dataset (SYnthetic NURsing) \cite{mediqa-synur-dataset} is a JSONL file split into \texttt{train}, \texttt{dev}, and \texttt{test}. Table \ref{tab:data_splits_singlecol} contains the dataset statistics of each split. Each instance contains a unique case identifier \texttt{id} and a free-text \texttt{transcript}. It also contains ground truth \texttt{observations}, represented as a list of objects from the schema with values that may be categorical (single- or multi-select), numeric, or free-text, reflecting the heterogeneity of routine nursing documentation.

\begin{table}[t]
\centering
\small
\begin{tabular}{lccc}
\hline
\hline
\textbf{Split} & \textbf{No. of Inst.} & \textbf{Total Obs.} & \textbf{Unique Obs.} \\
\hline
Train & 122 & 1685 & 166 \\
Dev & 101 & 1315 & 170 \\
Test & 199 & 2552 & 164 \\
\hline
\hline
\end{tabular}
\caption{Dataset statistics per split with the number of instances, total number of observations, and the number of unique observations in each split.}
\label{tab:data_splits_singlecol}
\end{table}


Across all splits, conversational nurse-patient transcripts are moderately long, averaging 192 words, with lengths ranging from 59 to 343 words. In the train and development sets, each instance contains 13.45 observations on average, ranging from 6 to 34 observations per case, with no duplicated concept IDs within an instance. Label frequency follows a long-tailed distribution, while a small set of observations occur very frequently (e.g., \emph{Cognitive status}, \emph{Mobility}, \emph{Oxygen saturation}, \emph{Nausea}), which are normally the most common observations made by nurses. Many concepts are sparse, with 55 concepts appearing at most five times and 23 concepts appearing only once or twice in the labeled data (e.g., \emph{Drain output}, \emph{Prosthetic use}, \emph{Foreign object removal}). 

The provided schema defines 193 unique observation concepts with explicit value types and, for categorical concepts, enumerated allowable values. The schema is dominated by categorical fields (130 \textsc{single\_select} and 12 \textsc{multi\_select}), with additional \textsc{numeric} (20) and \textsc{string} (31) concepts. Categorical concepts have relatively small label sets on average, but include larger option lists for clinically richer fields such as \emph{Bowel movement description} (12 options) and \emph{Pain severity} (11 options). The schema also includes 15 explicit ``unit'' concepts (e.g., oxygen saturation unit, respiration unit), representing measurements as value–unit pairs. In the labeled data, 178 out of the 193 schema concepts appear at least once, which supports broad coverage while preserving a realistic distribution.

Overall, this dataset is well-suited for evaluating schema-constrained extraction because it combines (i) conversational nurse-patient transcripts of varying length, (ii) a large and heterogeneous schema with strict type and enumeration requirements, including unit pairing, and (iii) a naturally imbalanced concept distribution that makes rare observations harder to capture while maintaining strong schema-adherence.

\section{Methodology}

\subsection{Overview}
Our approach is implemented as a modular workflow in which each component can be enabled or disabled, allowing controlled comparisons across design choices and LLM backbones. Figure \ref{fig.1} provides a complete illustration of our system. Given an input transcript $x$ and the schema $\mathcal{S}$, the system constructs a prompt that conditions the LLM on (i) task instructions enforcing strict structured output, (ii) schema (full or pruned schema), and (iii) retrieved in-context exemplars when RAG is enabled. The LLM is instructed to output only a JSON list of $(id, v)$ pairs (concept identifier and value), which we then deterministically expand to the required submission format by looking up each concept’s canonical \texttt{name} and \texttt{value\_type} from the schema; this avoids errors from hallucinated names or types and guarantees metadata adheres to the schema. Concretely, each final predicted observation is represented as \texttt{{id, name, value\_type, value}}, where \texttt{name} and \texttt{value\_type} are taken from $\mathcal{S}$ and \texttt{value} is generated by the LLM and validated against $\mathcal{S}$. We evaluate the same pipeline configurations with two LLM backbones: \textit{Llama-4-Scout-17B-16E-Instruct} (open-weights) \cite{meta2025llama} and \textit{GPT-5.2} (closed-weights) \cite{openai_gpt52_2025}, and report the comparative effects of RAG, schema-constraints (full or pruned schema), and second-pass auditing in the ablation study.

\subsection{Retrieval Corpus and Knowledge Base}
We use the training split as the sole retrieval corpus and source of in-context demonstrations. Each training instance is parsed from JSONL as a tuple $(x_i, y_i)$ containing a transcript $x_i$ and a gold label observation list $y_i$. We convert the gold label observations into a standardized representation $y_i^{\mathrm{std}} = [(id_j, v_j)]_{j=1}^{K_i}$, where $id_j$ is the concept identifier and $v_j$ is the corresponding value. We also derive a summary text for each training instance by mapping their gold label concept IDs to their concept names and concatenating them with a delimiter (e.g., ``name$_1$ $\mid$ name$_2$ $\mid \cdots$''). This yields two complementary textual views per training example: the original transcript text and a concept name summary text derived from the schema.

For semantic retrieval, we embed training instances and queries using the same encoder within each pipeline: Sentence-Transformers \texttt{all-MiniLM-L6-v2} \cite{hf_all_minilm_l6_v2} for the Llama-4 and OpenAI \texttt{text-embedding-3-large} \cite{openai_text_embedding_3_large} for the GPT-5.2. We pre-compute and cache embeddings for each transcript view (and, when retrieval is enabled, for the concept name view as well), along with $y_i^{\mathrm{std}}$ for use as few-shot examples. At inference time, the query transcript is embedded under the same encoder and compared against the cached training embeddings to retrieve candidate exemplars.

\subsection{Exemplar Retrieval for RAG}
When retrieval is enabled, we use an exemplar retrieval strategy that leverages both the narrative content of the transcript and the structured observation concepts associated with each training instance. Each training example is represented by two textual views: (i) the original transcript and (ii) a schema-derived concept-name summary constructed by mapping its gold label concept IDs to their standardized schema names. For a query transcript, we first retrieve a candidate pool using cosine similarity over transcript embeddings, and then re-rank the pool using a weighted score:
\[
\text{score(candidate pool)} = w_t. s_t + w_c. s_c + w_l. s_l
\]
where $s_t$ is cosine similarity between transcript embeddings, $s_c$ is cosine similarity between concept-name summary embeddings, and $s_l$ is a lexical overlap score computed as Jaccard similarity over token sets \cite{schutze2008introduction}. We set $(w_t, w_c, w_l) = (0.70, 0.25, 0.05)$ based on experiments on the dev set, as this combination yielded the best overall retrieval quality and downstream extraction performance. The top-$k$ exemplars under this score are inserted into the prompt as few-shot demonstrations, each consisting of the exemplar transcript paired with its standard gold label output as a JSON list of $(id, v)$ pairs. We vary the number of retrieved exemplars with $k \in \{3,5,10,20,30,50\}$ and analyze its effect.

\subsection{Schema Constraints}
We study two strategies for conditioning the LLM on the task schema $\mathcal{S}$, which enumerates all observation concepts and their value constraints. In \emph{full schema} prompting, we provide the complete schema to the model, including each concept’s identifier, standard name, value type, and (when categorical) its allowable value set. In \emph{pruned candidate schema} prompting, we instead construct a per-instance candidate concept set $\mathcal{C}(x)\subseteq \mathcal{S}$ intended to cover concepts relevant to transcript $x$ while limiting the output space. The prompt then includes only a compact candidate table for $\mathcal{C}(x)$ (identifier, name, value type, and truncated allowable values for categorical concepts).

To construct $\mathcal{C}(x)$, we score concepts by lexical match between transcript tokens and concept-name tokens (and, for categorical concepts, limited overlap with enumerated tokens), retaining the highest-scoring lexical candidates. We augment this with a semantic schema match by retrieving nearest concept names from an embedding index over all names in the schema and keeping matches above a fixed similarity threshold. We then expand $\mathcal{C}(x)$ using retrieval expansion by adding concept IDs observed in the retrieved exemplars’ gold label outputs, and we additionally inject a small set of common observation patterns (e.g., vital signs and frequently occurring nursing assessment fields). Finally, we size the candidate set to a target budget (with minimum and maximum bounds) by trimming low-scoring candidates while preserving a small set of common concepts, or padding with additional lexical/semantic candidates when the set is too small.

\begin{figure}[!t]
\begin{center}
\includegraphics[width=\columnwidth]{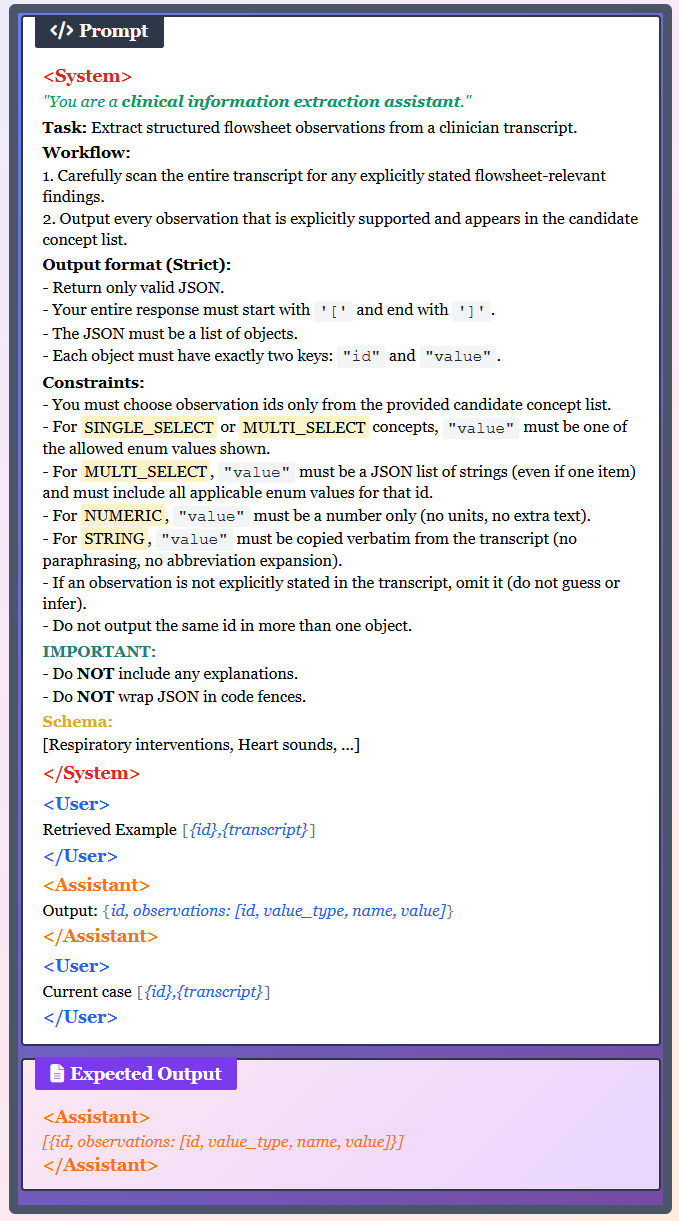}
\caption{The structured prompt with retrieved exemplars, schema, and the expected output.}
\label{fig.2}
\end{center}
\end{figure}

\begin{figure}[!ht]
\begin{center}
\includegraphics[width=\columnwidth]{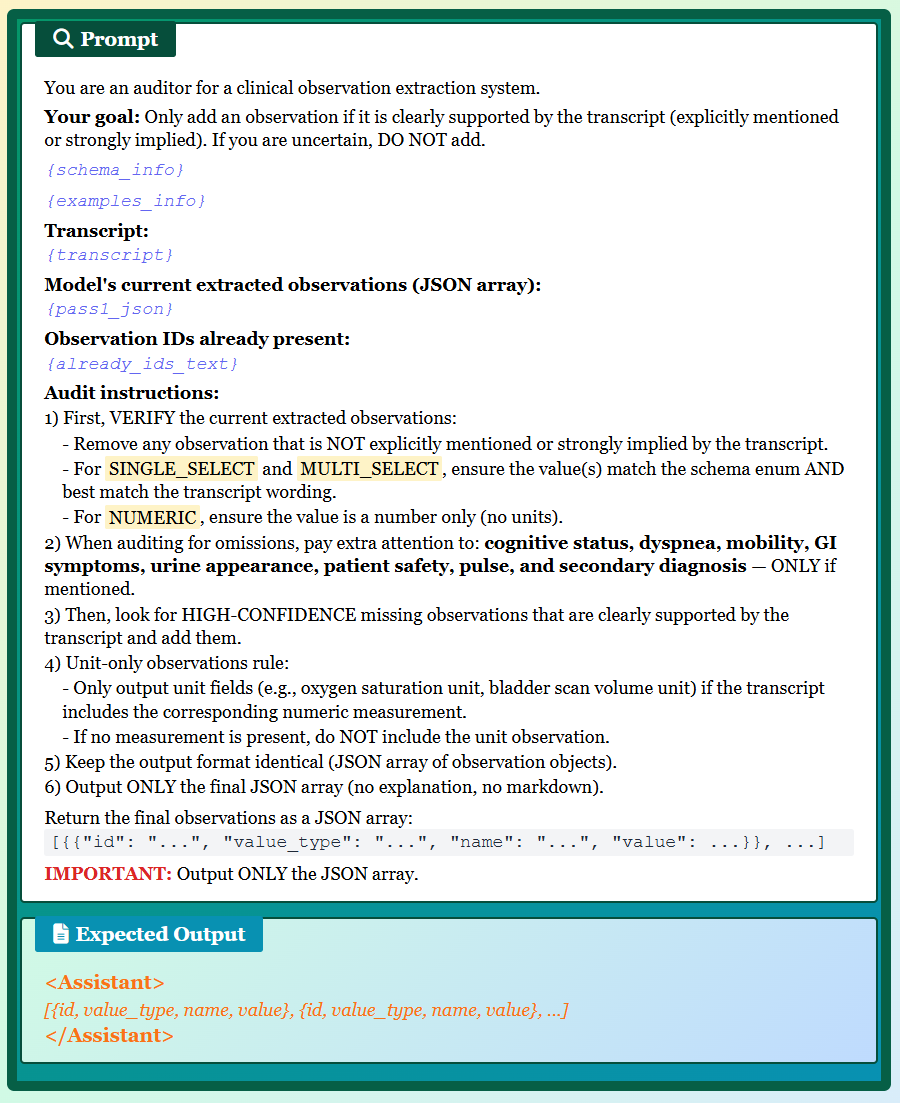}
\caption{The structured prompt for the second pass audit with retrieved exemplars, schema, first pass solution, and the expected output.}
\label{fig.3}
\end{center}
\end{figure}

\subsection{Prompt Construction and LLM Inference}
For each input transcript, we construct a prompt designed to elicit strictly structured outputs that adhere to the schema. Figure \ref{fig.2} illustrates the prompt, which begins with system-level instructions that enforce a JSON-only response and explicitly prohibit additional text, code fences, or explanations. We require a single JSON list of objects, each containing exactly an observation identifier and its value (i.e., $(id, v)$), and we instruct the model to omit any observation not supported by evidence in the transcript and to avoid duplicate identifiers.

We then provide either a full schema or a pruned candidate schema and add few-shot examples when RAG is enabled: up to $k$ retrieved exemplars formatted as a transcript followed by its gold label output. Finally, we append the query transcript and a concise instruction to produce the JSON output. We apply the same prompt template across LLM backbones, differing only in the inference interface (Transformers for \textit{Llama-4-Scout-17B-16E-Instruct} and API-based inference for \textit{GPT-5.2}).

\begin{table*}[t]
\centering
\renewcommand{\arraystretch}{1.15}
\begin{tabular}{l l r r r}
\hline
\hline
\textbf{LLM Backbone} & \textbf{Setting} & \textbf{Precision} & \textbf{Recall} & \textbf{$F_1$ score} \\
\hline
\hline
Llama-4 & Prompt-only + full schema & 72.92 & 52.84 & 61.28 \\
Llama-4 & Prompt-only + pruned candidate schema & 74.64 & 64.02 & 68.92 \\
Llama-4 & Prompt + RAG + full schema & 77.24 & 68.24 & 72.46 \\
Llama-4 & Prompt + RAG + pruned candidate schema & 78.58 & 68.50 & 73.20 \\
\hline
GPT-5.2 & Prompt-only + full schema & 81.97 & 73.17 & 77.32 \\
GPT-5.2 & Prompt-only + pruned candidate schema & 69.80 & 83.57 & 76.07 \\
GPT-5.2 & Prompt + RAG + full schema & 78.59 & 82.03 & 80.27 \\
GPT-5.2 & Prompt + RAG + pruned candidate schema & 81.71 & 73.54 & 77.41 \\
\hline
GPT-5.2 & Prompt + RAG + full schema + 2nd pass & 78.62 & 82.18 & \textbf{80.36} \\
\hline
\hline
\end{tabular}
\caption{Performance (\%) across Llama 4 and GPT-5.2 with schema/RAG configurations. Values are rounded to two decimals. Best $F_1$ is bolded.}
\label{tab:main_results}
\end{table*}

\subsection{Postprocessing and Schema Validation}
We apply postprocessing to make model outputs robust. First, we deterministically clean and parse the generated text into a JSON list of $(id, v)$ pairs, removing common formatting errors when necessary. We then validate predictions against the schema $\mathcal{S}$ by dropping unknown concept IDs and normalizing values according to each concept’s \texttt{value\_type}. For categorical concepts, outputs are constrained to allowable enumerated values; for multi-select concepts, values are converted to lists and deduplicated; and for numeric concepts, numeric strings are converted to numbers when possible. Duplicate IDs are removed by keeping the first valid occurrence. Finally, we expand each validated $(id, v)$ pair into the required submission format by retrieving the standard \texttt{name} and \texttt{value\_type} from the schema, producing \texttt{{id, name, value\_type, value}} entries.

\subsection{Second-Pass Auditing}
Figure \ref{fig.3} shows the prompt for the second pass auditing step. We evaluate this step as a refinement stage over the first pass, which is the initial system output produced by the LLM from a single pass over the transcript and schema. The auditor is prompted with the original transcript, the same schema used in the first pass (full schema or pruned candidate schema), and the first-pass prediction, and is instructed to remove unsupported items, correct schema adherence issues, and add only clearly supported missing observations. The audited output is then passed through the same schema validation and deterministic expansion step as the first pass.

\subsection{Evaluation Strategy}
We compute precision, recall, and $F_1$ over extracted observations by matching predicted and reference items by concept \texttt{id} and \texttt{value}. For \textsc{multi\_select} concepts, the evaluation script expands each selected value into a separate item prior to matching, effectively scoring multi-select outputs at the individual choice level. Categorical and string values are compared as strings, and numeric values are compared after conversion to floating point. 

\section{Results and Discussion}

\subsection{Overview}
Table~\ref{tab:main_results} reports the performance of our retrieval-augmented, schema-constrained extraction pipeline across LLM backbones and design variants. The strongest configuration uses \textit{GPT-5.2} with OpenAI \texttt{text-embedding-3-large} for exemplar retrieval, combined with full schema prompting and $top\text{-}k=30$ retrieved exemplars, yielding the best overall $F_1$. We observe that increasing the number of retrieved exemplars beyond this point produces only marginal changes, suggesting diminishing returns from additional exemplars. Finally, adding a second-pass auditing stage yields a small additional improvement, indicating that most performance gains are driven by the primary system, with second-pass primarily addressing residual normalization and schema-adherence issues.

\subsection{Ablation Study}
We conduct an ablation analysis to isolate the contributions of RAG, schema, and second-pass auditing. We start with a prompt-only baseline using \textit{Llama-4-Scout-17B-16E-Instruct} and the full schema, which illustrates the difficulty of extracting and normalizing concepts when the model must reason over a large output space under strict type and enumeration constraints. In this case, errors are dominated by missed concepts and schema-inconsistent outputs, reflecting both limited grounding from the input and the challenge of selecting among many semantically adjacent concepts.

Adding retrieval augmentation improves performance for both LLMs. Concretely, retrieved examples reduce false negatives by providing localized evidence for what constitutes an extractable observation and how it should be normalized under the schema. They also reduce formatting and validity errors by reinforcing the expected JSON structure and value through examples. This effect is especially visible for categorical concepts, where the examples encourage selecting values from the allowed set rather than generating free-form paraphrases.

We then study different schema strategies. For open-weight LLM, pruned candidate schema prompting produces the largest improvement, yielding the best results for \textit{Llama-4} when combined with retrieval. This behavior is consistent with a constrained-decision-space effect: pruning removes many implausible concepts, making it easier for a smaller model to adhere to the schema and to focus extraction on the most relevant portion of the schema. This change improves recall while keeping precision stable or slightly improved, indicating that the model misses fewer relevant observations without a corresponding increase in over-extraction.

For the larger close-weights model, we observe a different interaction between retrieval and schema-constraint. Candidate pruning tends to increase recall but decrease precision, consistent with a checklist-like bias in which the reduced candidate set implicitly encourages broader extraction even when supporting evidence is weak. Under retrieval augmentation, the best \textit{GPT-5.2} performance is instead obtained with full schema prompting, suggesting that a larger model can exploit entire schema context to better disambiguate related concepts while using retrieved exemplars primarily for grounding and normalization. Finally, adding a second-pass auditing stage on top of the best \textit{GPT-5.2} configuration yields a small additional gain. The magnitude of this improvement indicates that most performance is already captured by the combination of retrieval and schema-constraints, with the second-pass primarily correcting residual schema-adherence and normalization errors rather than changing extraction coverage.

Taken together, the ablation results support a clear finding: retrieval augmentation is broadly beneficial, but the optimal schema constraint is model-aware. Pruned candidate schema prompting is critical for controlling the effective output space for the smaller open-weight model, whereas the larger closed model benefits more from full schema context when paired with RAG. The best system performance is because of the combination of (i) exemplar grounding that reduces omissions and stabilizes normalization, and (ii) a full schema that enables consistent disambiguation under strict type and enumeration requirements.

\begin{table}[h]
\centering
\renewcommand{\arraystretch}{1.15}
\small
\setlength{\tabcolsep}{6pt}
\begin{tabular}{l r}
\hline
\hline
\textbf{Error type} & \textbf{Count}\\
\hline
\textbf{False positives} & \textbf{608}\\
\quad Spurious concept (id not in gold) & 498\\
\quad Wrong value (id in gold) & 110\\
\hline
\textbf{False negatives} & \textbf{485}\\
\quad Missed concept (id not predicted) & 382\\
\quad Wrong value (id predicted) & 103\\
\hline
\hline
\end{tabular}
\caption{Breakdown of false positives (FP) and false negatives (FN) into spurious concept errors versus value mismatches.}
\label{tab:error_breakdown}
\end{table}

\section{Error Analysis}
We analyze errors by comparing predicted observations against the reference output (with \textsc{multi\_select} values treated as one item per selected choice). Across the test set (199 cases), the system produces 2236 true positives, 608 false positives, and 485 false negatives, as shown in Table \ref{tab:error_breakdown}, indicating that most residual errors arise from either predicting extra items or missing gold label items rather than near-miss value disagreements. Consistent with this, 81.9\% of false positives (498/608) correspond to \emph{spurious predictions} whose concept IDs are not present anywhere in the ground truth set for that case, while the remaining 18.1\% (110/608) are cases where the correct concept is predicted but with an incorrect value. On the false-negative side, 78.8\% (382/485) are \emph{omitted concepts} (the concept ID is not predicted at all), and 21.2\% (103/485) are attributable to value mismatches. We also find that errors increase primarily with \emph{clinical content density}: the number of errors correlates more strongly with the number of gold label items ($\rho=0.45$) and the number of predicted items ($\rho=0.53$) than with transcript length ($\rho=0.24$), suggesting that dense dictations amplify both omission risk and over-extraction.

\begin{table*}[htb]
\centering
\renewcommand{\arraystretch}{1.15}
\small
\setlength{\tabcolsep}{3.5pt}
\begin{tabular}{r l l r |r l l r}
\hline
\hline
\multicolumn{4}{c}{\textbf{Top FP concepts}} & \multicolumn{4}{c}{\textbf{Top FN concepts}}\\
\hline
\textbf{id} & \textbf{Name} & \textbf{Type} & \textbf{FP} &
\textbf{id} & \textbf{Name} & \textbf{Type} & \textbf{FN}\\
\hline
3   & Respiratory interventions & \textsc{multi\_select} & 65 &
3  & Respiratory interventions & \textsc{multi\_select} & 73\\
31  & Secondary diagnosis & \textsc{string} & 61 &
148 & Gastrointestinal symptoms & \textsc{multi\_select} & 29\\
0   & Broset violence checklist & \textsc{single\_select} & 41 &
0  & Broset violence checklist & \textsc{single\_select} & 23\\
162 & Patient identification & \textsc{string} & 26 &
6  & Weightbearing status & \textsc{multi\_select} & 20\\
6   & Weightbearing status & \textsc{multi\_select} & 19 &
26  & Delirium symptoms & \textsc{string} & 17\\
26  & Delirium symptoms & \textsc{string} & 17 &
7  & Oral mucosa status & \textsc{single\_select} & 16\\
7   & Oral mucosa status & \textsc{single\_select} & 17 &
179 & Temperature unit & \textsc{single\_select} & 15\\
117 & Patient safety & \textsc{single\_select} & 15 &
89  & Mobility & \textsc{single\_select} & 15\\
130 & Cognitive status & \textsc{single\_select} & 13 &
107 & Fall risk identification & \textsc{single\_select} & 14\\
67  & Dyspnea & \textsc{single\_select} & 12 &
130 & Cognitive status & \textsc{single\_select} & 14\\
\hline
\hline
\end{tabular}
\caption{Most frequent concept-level errors. False Positive (FP) counts reflect extra predicted items; False Negative (FN) counts reflect missing gold-labeled items (after expanding \textsc{multi\_select} choices).}
\label{tab:top_error_concepts}
\end{table*}

A notable fraction of errors concentrates in a small set of systematic patterns. First, we observe a persistent class of representation-level mismatches involving a small subset of low-number concept identifiers that appear with leading zeros in the reference output (e.g., \texttt{03} vs.\ \texttt{3}). These mismatches produce both false positives and false negatives even when the extracted content is clinically correct, and they account for a substantial share of total errors (147/485 false negatives and 159/608 false positives). A related phenomenon occurs for temperature units, where the schema’s categorical tokens do not match the reference encoding, yielding systematic value mismatches. These issues motivate future work on \emph{robust standardization} for heterogeneous schema and annotations, including normalization of identifier formats and encoding-aware mapping for categorical tokens, as a general requirement when deploying schema-constrained extraction across data sources with inconsistent conventions.

Second, the largest precision losses arise from a small number of \emph{over-predicted} concepts that are plausible given the transcript but rarely used in the references. For example, Table \ref{tab:top_error_concepts} shows that \texttt{Secondary diagnosis} appears only 3 times in the gold labels but is predicted 61 times, and \texttt{Patient identification} appears only 3 times in the gold labels but is predicted 26 times. This pattern suggests that the model often identifies clinically relevant information but maps it into schema slots that are sparsely annotated or used under stricter, dataset-specific criteria. Addressing this reliably is not a simple rule-based fix; it requires learning concept-specific decision policies and calibration that reflect when a mention should be normalized into a structured field versus left unexpressed.

Finally, residual recall errors frequently reflect \emph{implicit or distributed evidence} that is harder to consolidate into a single normalized observation, especially for checklist-style multi-select fields (e.g., \texttt{Gastrointestinal symptoms}) and summary concepts (e.g., \texttt{Mobility}). In addition, free-text fields exhibit high brittleness under strict value matching: \texttt{Pain description} is present 12 times in gold labels and is frequently predicted, yet none of these predictions match exactly at the string level, indicating that minor paraphrasing or formatting differences can dominate the residual error mass. Improving these cases likely requires methods that better ground string-valued slots in transcript spans (e.g., structured span-copying or alignment) and that model schema overlap explicitly so that evidence expressed in supporting fields is more consistently propagated into the corresponding summary concepts.

\section{Conclusion}
Schema-constrained extraction from conversational nurse-patient transcripts is an important step toward reducing documentation burden by transforming clinical documentation into structured, schema-adherent observations. Our work presents a modular retrieval-augmented extraction pipeline that combines training-set exemplar retrieval, schema-constrained prompting (full schema or pruned candidate schema), deterministic schema-based postprocessing, and a second-pass audit to refine outputs. Our results support three key findings: pruned candidate schema is beneficial for smaller open-weight models but can be counterproductive for larger models that instead benefit from richer full schema context; retrieval-augmented generation consistently improves performance, with gains strongly influenced by how schema information is presented; and second-pass auditing provides modest additional improvements by correcting residual schema-adherence and normalization issues. Overall, these findings underscore that effective schema-constrained observation extraction is achieved by jointly selecting retrieval and schema-constraint strategies that match the capabilities of the underlying model.










\section{References}\label{sec:reference}

\bibliographystyle{lrec2026-natbib}
\bibliography{lrec2026-example}


\end{document}